\documentclass[10pt,a4paper,onecolumn,oneside]{article}
\pdfoutput=1
\usepackage{subfig}
\usepackage{cite}
\usepackage{amsmath,amssymb,amsfonts}
\usepackage{algorithmic}
\usepackage{graphicx}
\usepackage{textcomp}
\usepackage{xcolor}
\usepackage{multicol}
\def\BibTeX{{\rm B\kern-.05em{\sc i\kern-.025em b}\kern-.08em
    T\kern-.1667em\lower.7ex\hbox{E}\kern-.125emX}}

\newcommand{\U}{\mathbf{A}}
\newcommand{\V}{\mathbf{B}}
\newcommand{\1}{\mathbf{1}}
\newcommand{\footremember}[2]{%
    \footnote{#2}
    \newcounter{#1}
    \setcounter{#1}{\value{footnote}}%
}
\newcommand{\footrecall}[1]{%
    \footnotemark[\value{#1}]%
} 
\newtheorem{lemmax}{\textbf{Lemma}}
\newtheorem{theoremx}{\textbf{Theorem}}
\newtheorem{definitionx}{\textbf{Definition}}

\title{Comparing Graph Clusterings: \\Set partition measures vs. Graph-aware measures}

\author{%
Val\'{e}rie Poulin\footremember{vpaff}{The Tutte Institute for Mathematics and Computing, Ottawa, ON, Canada.}%
  \and Fran\c{c}ois Th\'{e}berge\footrecall{vpaff}%
}

\begin{document}
\maketitle

\begin{abstract}
In this paper, we propose a family of graph partition similarity measures that take the topology of the graph into account.  These graph-aware measures are alternatives to using set partition similarity measures that are not specifically designed for graph partitions. The two types of measures, graph-aware and set partition measures, are shown to have opposite behaviors with respect to resolution issues and provide complementary information necessary to assess that two graph partitions are similar.

\end{abstract}

\section{Introduction}
\label{section:intro}
An impressive number of graph clustering algorithms have been proposed, studied and compared over the past decades \cite{girvan2002community, clauset2004finding, rosvall2007information, raghavan2007near, newman2006finding, louvain, reichardt2006statistical}. To identify better graph clustering techniques, one needs a way to score the techniques against one another. A typical method is to compare values of some similarity measure between ground truth partitions of given graphs and the partitions produced by the different algorithms on those graphs. However, the choice of the similarity measure used is crucial and has a huge impact on the conclusions made.

In graph clustering comparison studies \cite{lancichinetti2009community, yang2016comparative, orman2009comparison, fortunato2016community}, set partition similarities are used as accuracy measures. Typically, a member of the pair-counting family \cite{Albatineh2006,arabie} such as Adjusted Rand Index, or of the Shannon information-based family \cite{meila,vinh2010,vinh} such as Adjusted Mutual Information is used to assess the superiority of a graph clustering algorithm over another. These measures are designed for comparing set partitions and not graph partitions specifically. We call them {\it graph-agnostic} as they ignore the graph structure. 

In this paper, we propose a family of {\it graph-aware} measures for graph partition similarity with their adjusted forms. We compare the graph-aware with the graph-agnostic partition measures and demonstrate that the two types of measures offer complementary views of the typical clustering errors known as resolution errors and, therefore, should be used jointly before any proper conclusions can be made on the accuracy of a graph clustering algorithm. 

The paper is organized as follows: In Section~\ref{section:def}, we set the notation. Section~\ref{section:clustcomp} presents the most common and widely used families of set partition similarity measures. In Section~\ref{section:edgemeasure}, we define a family of graph-aware similarity measures, we prove a result on the complementarity of the two types of measures and we propose an adjustment for the family. Some experiments are presented in Section~\ref{section:both} to study the impact of the adjustments and to study the relation between graph-aware and graph-agnostic measures. 

\section{Notation}
\label{section:def}
We define $G=(V,E)$ an graph where $V=\{1,2,...,n\}$ is the set of vertices and $E \subset \{(x,y) | x,y\in V, x<y\}$, the set of edges. All graphs considered are undirected. We use $V(G)$ to refer to the vertices of G and $E(G)$ to refer to its edge set. If $A \subset V$, then $G_{A}$ denotes the subgraph of $G$ obtained by restricting $G$ to vertices in $A$.

Let $\U=\{A_1, \cdots, A_{k_a}\}$ and $\V=\{B_1, \cdots, B_{k_b}\}$ denote two partitions of $V$. The cardinality of the partitions $\U$ and $\V$ are $k_a$ and $k_b$ and the cardinalities of each of the parts are $|A_i| = a_i \,\mbox{ for } \, i=1,\cdots,k_a$ and $|B_j| = b_j \,\mbox{ for } \, j=1,\cdots,k_b$. Finally, the size of the pairwise intersections are $|A_i \cap B_j| = n_{ij}$.

\begin{definitionx}
$\U$ is a {\it connected partition of $G$} if $\U$ is a partition of $V$ and if all subgraphs $G_{A_i}$ are connected.
\end{definitionx}

\section{Graph-agnostic clustering comparison measures}
\label{section:clustcomp}
Similarity measures between {\it set} partitions have been well studied \cite{Albatineh2006,arabie,vinh,meila2005}. The most widely used similarity measures lie in one of the two following families: pair-counting ($PC$) measures and mutual information ($MI$) based measures. In this section, we define the two families.

Let $\displaystyle{ P_\U = \cup_{i=1}^r \{(x,y) \in A_i \times A_i \, | \, x<y \}}$ denote the pairs of points lying in the same part of $\U$.
We define $P_\V$ similarly, and we use an overline to denote the complement of a set: $\overline{P_\U} = \{ (x,y) \in (V \times V) \, | \, x<y \mbox{ and } (x,y) \notin P_\U \}$. The two  pair counting indices that were first proposed are the Rand Index ($R$) \cite{rand1971} and the Jaccard Index ($J$):
$$\displaystyle{R(\U,\V) = \frac{|P_\U \cap P_\V| + |\overline{P_\U} \cap \overline{P_\V}|}{\binom{n}{2}}} \mbox{, }\displaystyle{J(\U,\V) = \frac{|P_\U \cap P_\V|}{|P_\U \cup P_\V|}}.$$ 

The key value of most pair counting similarity measures is $|P_\U \cap P_\V|$, the number of pairs belonging to the same parts in both partitions. The Rand index is an exception as it also includes the number of pairs belonging to different parts in both partitions. Different normalizations are used to ensure the measures are constraint to values in $[0,1]$. Other members of the family of pair counting similarity measures are:
$$PC_f(\U,\V) = \frac{|P_\U \cap P_\V|}{f(|P_\U|, |P_\V|)},$$
where $f \in \{mn, gm, min, max\}$ which denote the mean, geometric mean, minimum or maximum function respectively.

Another family of measures used for measuring the similarity of partitions is the Shannon information based family. 
The entropy of a partition $\U$ is defined as 
$H(\U) = -\sum \frac{a_i}{n}\log \frac{a_i}{n}$, the joint entropy
of $\U$ and $\V$ as $H(\U,\V) = - \sum_{i,j}\frac{n_{ij}}{n} \log \frac{n_{ij}}{n}$, and, the mutual information
between two partitions is $I(\U,\V) = \sum_{i,j} \frac{n_{ij}}{n} \log \frac{n_{ij}/n}{a_ib_j/n^2}$.
The mutual information between two partitions is the core value for comparing partitions
with information-based measures. Similarly to the pair counting members, information based measures are normalized versions of the mutual information having the unit interval as image:
$$MI_f(\U,\V) = \frac{I(\U,\V)}{f(H(\U), H(\V))},$$
where $f$ is as for $PC_f$. In \cite{vinh2010}, they show that $MI_{max}$ is a true metric and argue that it should be favored over the other measures. 

The two families of measures suffer from the problem of not having a constant baseline of 0 when the compared partitions are random and independent. For this reason, adjusted forms were proposed independently for the pair counting \cite{arabie} and information-based \cite{vinh} families. The adjustments consist of subtracting the expected value of the measure under a random model namely, the {\it permutation model}. The permutation model consists of the expected measure between random partitions $\U$ and $\V$ given $a_i$ and $b_j$ their marginals\footnote{The choice of random model is discussed in \cite{gates2017impact}.}. The expectation values can be obtained empirically but closed forms exist for all but the Jaccard measure. Some measures collapse to having the same adjusted forms as in the case for the Rand Index and the $PC_{mn}$. The adjusted Rand Index (ARI) is:
$$\mathit{ARI}(\U,\V) = \frac{|P_\U \cap P_\V| - |P_\U||P_\V|/\binom{n}{2}}{\frac{1}{2}(|P_\U|+|P_\V|) - |P_\U||P_\V|/\binom{n}{2}}.$$ 
Note that what is called the {\it Adjusted Mutual Information}, $\mathit{AMI}(\U,\V)$, is the adjusted form of the measure $\mathit{MI}_{max}$. 

In \cite{romano2016adjusting}, a broader family based on the generalized Tsallis $q$-entropy is proposed which unifies the two families. The authors analytically compute adjustments for this family which generalizes the adjustments that were derived independently for the two families. As a consequence of their work, it is justified to use one of $AMI$ or $ARI$ to compare the accuracy of clustering algorithms given a ground truth partition if the number of data points is large relative to the number of parts in the partition.

\section{Graph-aware clustering comparison measures}
\label{section:edgemeasure}
\begin{figure}[h]
\begin{center}
\centering
\includegraphics[angle=0,width=.7\linewidth]{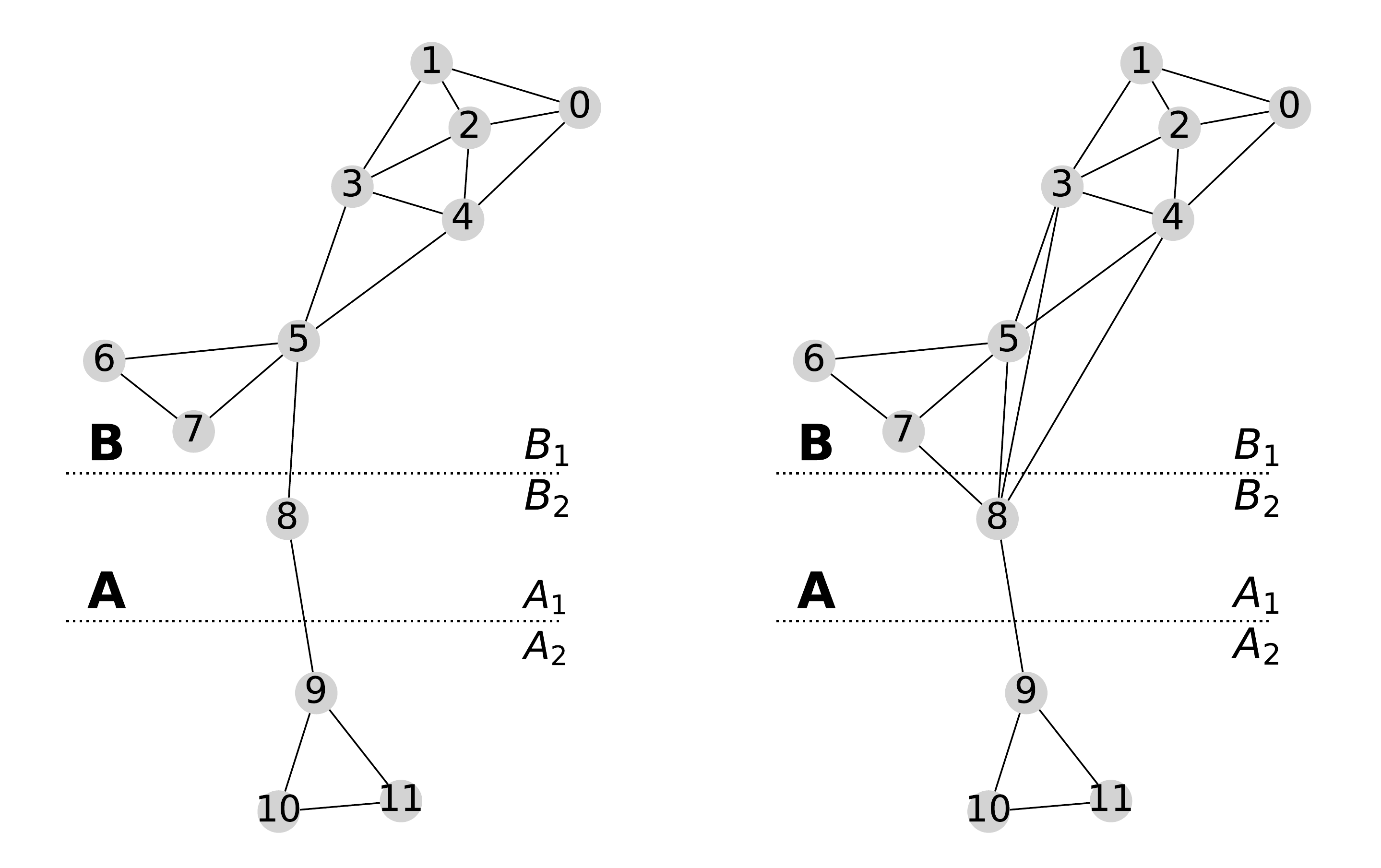} 
\caption{\small Partitions $\U$ and $\V$ are identical as set partitions on $G_1$ and $G_2$, and so their similarities are the same when using graph-agnostic measures regardless of the underlying graphs.} 
\label{fig:graph_vs_set}
\end{center}
\end{figure}

The clustering comparison measures discussed in the previous section account for the nodes in a graph but ignore the edges. Should the similarity between partitions $\U$ and $\V$, shown in Figure~\ref{fig:graph_vs_set}, be the 
same on graph $G_1$ as on graph $G_2$? When restricting to vertices, the two cases, $G_1$ and $G_2$, are indeed 
identical. However, when including the edges, the impact of placing vertex 8 in one part or the other is quite different on both graphs. 
In this section, we introduce a family of similarity measures for graph partitions that take edges into account and prove that both graph-agnostic and graph-aware measures are critical for effectively comparing graph partitions.

\subsection{Non-adjusted graph-aware similarity measures}

\noindent A connected graph partition $\U$ can be described in two ways. The most intuitive one is to
consider $\U$ as a partition of the graph's vertices, as we did in Section~\ref{section:clustcomp}.
Another way, is with a binary classification of the edges: the endpoints of the edges being either in the {\it same part} or in {\it different parts} of the partition.
Hence, a connected partition $\U$ of a graph $G$ induces a binary edge classification:
\begin{eqnarray*}
b_\U: E(G) &\rightarrow& \{0,1\} \\
(i,j) &\mapsto& \left\{
\begin{array}{ll}
1 & \quad \exists A_r \in \U \, \mbox{ with } \, i,j \in A_r, \\
0 & \quad \mbox{otherwise.} 
\end{array}
\right.
\end{eqnarray*}
Not all binary edge classifications correspond to a connected graph partition. However, the set of binary edge classifications $\{0,1\}^{E}$, where $E=E(G)$, can be grouped into equivalence classes w.r.t. the graph partition they induce.

\begin{definitionx}
Let $b \in \{0,1\}^{E}$ and consider $G' = (V, b^{-1}(1))$ a subgraph of $G$ formed of all class-one edges of $b$. We say $G'$ is the subgraph {\it induced by the classification} $b$.

\noindent Let $b_1, b_2 \in \{0,1\}^{E}$. The binary classification $b_1$ {\it is in relation} with $b_2$ if the two subgraphs induced by the classifications have identical connected components on $G$. In that case, we write $b_1 \equiv_{G} b_2$. 
\end{definitionx}

\begin{figure}[ht]
\begin{center}
\centering
\includegraphics[angle=0, width=.8\linewidth]{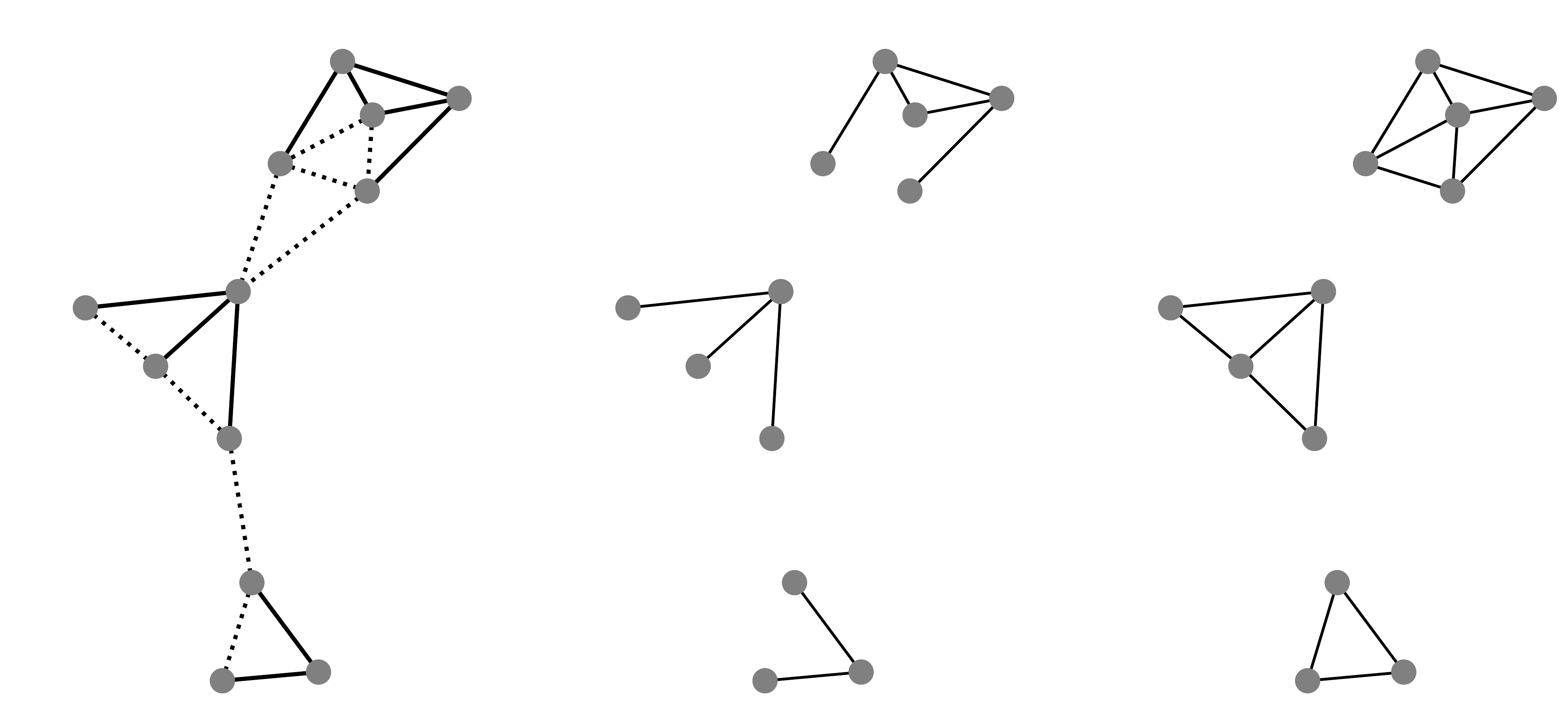} 
\caption{\small From left to right: edge classification $b$, the class-1 edges corresponding to $b$, and $b$'s class representative $\overline{b}^{_G}$ which includes all edges of the connected graph partition. } 
\label{fig:class_rep}
\end{center}
\end{figure}

The quotient set $\{0,1\}^{E} / \equiv_{G}$ divides the set of binary edge classifications into equivalence classes $[b]$ where members of a class all induce the same connected partition on $G$. For each class $[b]$ we define its {\it representative class member}, $\overline{b}^{_G}$, as the binary classification having the largest number of class-one edges in that class: $\overline{b}^{_G}(i,j)
 = \max\{b(i,j):b \in [b]\}.$ For each $b \in \{0,1\}^{E}$, there exists a connected graph partition $\U$ such that $\overline{b}^{_G} = b_\U$. Clearly, if $\U$ is a connected partition of $G$, $\overline{b}_\U^{_G} = b_\U$. See Figure~\ref{fig:class_rep} for an illustration. 

This partition description opens the door to new ways of measuring the similarity between two graph partitions using the corresponding edge classifications. With a fixed arbitrary ordering of the edges, the binary classifications are considered as binary vectors in $\{0,1\}^{|E|}$. We use $|b|$ to denote the L1-norm of the vectors and so expressions such as $|b^{-1}(1)|$ and $|b_1^{-1}(1) \cap b_2^{-1}(1)|$ are replaced by $|b|$ and $|b_1 \cdot b_2|$. 

Given two binary edge classifications $b_1, b_2 \in \{0,1\}^{|E|}$, we define their similarity as
$$S_G(b_1, b_2) = S^*(\overline{b}_1^{_G}, \overline{b}_2^{_G}) = S^*(b_\U, b_\V),$$
where $S^*$ is a pre-determined similarity measure on binary vectors and $\U$ and $\V$ are the two connected graph partitions associated to $\overline{b}_1^{_G}$ and $\overline{b}_2^{_G}$. There are four core values on which all binary similarity measures are based: $a_{00}, a_{01}, a_{10}$ and $a_{11}$ where $a_{ij}$ denotes the number of elements on which  $b_\U$ takes value $i$ and $b_\V$ takes value $j$: $a_{11}=|b_\U \cdot b_\V|$, $a_{01}=|(\1-b_\U) \cdot b_\V|$, $a_{10}=|b_\U \cdot (\1-b_\V)|$ and $a_{00}=|(\1-b_\U) \cdot (\1-b_\V)|$, where $\1=(1,1,\cdots,1)$. In the graph context, those four values can be expressed as: 
{\renewcommand{\arraystretch}{1.2}
\begin{center}
\begin{tabular}{c| c | c}
$b_\U$/$b_\V$ & 1 & 0 \\
\hline
1 & $\displaystyle{a_{11} =|P_\U \cap P_\V \cap E|}$ &  $\displaystyle{a_{10} = |P_\U \cap \overline{P_\V} \cap E|}$ \\
0 & $\displaystyle{a_{01} = |\overline{P_\U} \cap P_\V \cap E|}$ &  $\displaystyle{a_{00} = |\overline{P_\U} \cap \overline{P_\V} \cap E|}$ 
\end{tabular}
\end{center}
}
\noindent 
Well-known binary similarity measures include accuracy, the F-score, Cosine similarity, Simpson, Braun \& Banquet, etc. See \cite{lapalme, Cha} for an exhaustive list of measures. It turns out that those measures relate to the pair counting measures
described in Section~\ref{section:clustcomp}, obtained by restricting the counts to pairs
of vertices sharing an edge, see Table~\ref{tab:gpc}.
\begin{table*}[t]
\caption{\small From classification measures to graph-aware clustering measures.}
\label{tab:gpc}
\begin{center}
\footnotesize{
\begin{tabular}{lll}
Accuracy  & $\displaystyle{\frac{a_{00}+a_{11}}{\sum_{i,j}a_{ij}}}$  & $RI(\cdot;G):  \displaystyle{\frac{|P_\U \cap P_\V \cap E| + |\overline{P_\U} \cap \overline{P_\V} \cap E|}{|E|}}$ \\
F-score ($\beta=1$)  & $\displaystyle{\frac{2a_{00}}{2a_{00}+a_{01}+a_{10}}}$  & $\mbox{PC}_{mn}(\cdot;G)$:  $\displaystyle{\frac{|P_\U \cap P_\V \cap E|}{\frac{1}{2}(|P_\U \cap E| + |P_\V \cap E|)}}$ \\
Cosine   & $\displaystyle{\frac{a_{00}}{\sqrt{(a_{00}+a_{10})(a_{00}+a_{01})}}}$  & $\mbox{PC}_{gmn}(\cdot;G)$:  $\displaystyle{\frac{|P_\U \cap P_\V \cap E|}{\sqrt{|P_\U \cap E| |P_\V \cap E|}}}$ \\
Simpson   & $\displaystyle{\frac{a_{00}}{\min\{ (a_{00}+a_{10}),(a_{00}+a_{01})\}}}$  & $\mbox{PC}_{min}(\cdot;G)$:  $\displaystyle{\frac{|P_\U \cap P_\V \cap E|}{\min \{|P_\U \cap E| , |P_\V \cap E|\}}}$ \\
Braun\&Banquet   & $\displaystyle{\frac{a_{00}}{\max\{ (a_{00}+a_{10}),(a_{00}+a_{01})\}}}$  & $\mbox{PC}_{max}(\cdot;G)$:  $\displaystyle{\frac{|P_\U \cap P_\V \cap E|}{\max \{ |P_\U \cap E| , |P_\V \cap E| \}}}$ \\
\end{tabular}
} 
\end{center}
\end{table*}

\subsection{Properties of graph-aware and graph-agnostic measures}

Different algorithms produce partitions of different sizes and many are known to suffer from the resolution issue \cite{Fortunato36,Kumpula2007}. It is therefore interesting to understand how the measures behave on partitions of various resolutions. 

A partition $\V$ is said to be a {\it refinement} of a partition $\U$, denoted $\V < \U$, if each part of $\V$ is a subset of a part of $\U$. In that case, we also say that $\U$ is a {\it coarsening} of $\V$.
The following result demonstrates that the graph-aware and the non-adjusted pair-counting measures behave differently with respect to partition refinements or coarsenings if the underlying graph has some community structure.  We will use $\mathcal{G}(n,k_1,k_2,\U)$, a variant of Girvan and Newman model \cite{girvan2002community, condon2001algorithms} to study a simple family of graphs having community structure. Graphs in $\mathcal{G}(n,k_1,k_2,\U)$ have $n$ vertices split into a partition $\U$: $k_1$ edges are randomly placed between pairs of vertices in {\it same} parts of $\U$ and $k_2$ are randomly placed between pairs of vertices in {\it different} parts, $k_1\leq |P_\U|$ and $k_2 \leq |\overline{P_\U}|$. Note that with this random process, $\U$ is not necessarily a connected partition of the random graphs. Let $p=k_1/|P_\U|$ and $q=k_2/|\overline{P_\U}|$, to simplify the notation, we write $G_\U \sim \mathcal{G}(n,p,q,\U)$ to denote $G_\U \sim \mathcal{G}(n,k_1=p|P_\U|,k_2=q|\overline{P_\U}|,\U)$. Moreover, we write $G_\U$ to emphasize the fact that there is an underlying partition $\U$ in the random generation of the graph. 
\begin{lemmax}
Consider $G_\U \sim \mathcal{G}(n,p,q,\U)$ with $\V_1 > \U$ a coarsening of $\U$ and $\V_2 < \U$, a refinement of $\U$. Then
\begin{itemize}

\item[(i)] $\mathbb{E}_{G_\U}[PC_{mn}(\U,\V_1;G_\U)] \geq PC_{mn}(\U,\V_1)$, 
if $p \geq q$.

\item[(ii)] $\mathbb{E}_{G_\U}[PC_{mn}(\U,\V_2;G_\U)] \leq PC_{mn}(\U,\V_2)$, for all $p$, $q$ values.

\end{itemize}
\label{lem}
\end{lemmax}

{\it proof} 
Let $a = |P_\U|$, $x_1 = |P_{\V_1} \backslash P_\U|$, $x_2 = |P_{\U} \backslash P_{\V_2}|$, $X_1 = |P_{\V_1} \backslash P_\U \cap E|$ and $X_2 = |P_{\U} \backslash P_{\V_2} \cap E|$. $X_1$ and $X_2$ are two independent hypergeometric random variables: $X_1 \sim Hyper(x_1, q, a)$ and $X_2 \sim Hyper(x_2, p, a)$ with $\mathbb{E}(X_1)=qx_1$ and $\mathbb{E}(X_2)=px_2$. Note that $|P_\U \cap E| = pa$, $|P_{\V_1} \cap E| = pa + X_1$, $|P_{\V_2} \cap E| = pa - X_2$, $P_\U \cap P_{\V_1} = P_\U$ and $P_\U \cap P_{\V_2} = P_{\V_2}$. 
\begin{itemize}
\item[(i)] Let $Z = \frac{pa}{pa + 1/2 X_1}$. Since, $Z > 0$, $1/Z$ is a convex function of $Z$. We have 
$$
\mathbb{E}(Z) \geq \left[ \mathbb{E}(1/Z) \right]^{-1} \\
= \left[ \frac{pa+1/2qx_1}{pu} \right]^{-1} \\
\geq PC_{mn}(\U, \V_1), \mbox{ if } p \geq q.
$$
\item[(ii)] Again, we use the convex function trick with $Z=\frac{pa}{pa-1/2X_2}$, $Z > 0$. 
\end{itemize}
$$
\mathbb{E}\left( \frac{pa-X_2}{pa-1/2X_2}\right) = 2 - \mathbb{E}(Z) \\
\leq 2 - \left[ \frac{a-1/2x_2}{a} \right]^{-1} \\
= PC_{mn}(\U,\V_2).  
$$

This lemma shows how the graph-aware and graph-agnostic similarity measures compare to one another given refinements or coarsenings of the ground truth partition of a graph. The following result is very important for understanding the degradation of similarities given different types of perturbations (groupings or splittings) of the ground truth partition. 
\begin{theoremx}
Consider $G_\U \sim \mathcal{G}(n,p,q,\U)$ with $\V_1 > \U$ a coarsening of $\U$ and $\V_2 < \U$, a refinement of $\U$ such that $|P_\U|^2 < |P_{\V_1}| \cdot |P_{\V_2}|$. Then
\begin{itemize}

\item[(i)]  $PC_{mn}(\U,\V_1) < PC_{mn}(\U,\V_2)$.

\item[(ii)]  
$\mathbb{E}_{G_\U}[PC_{mn}(\U,\V_1;G_\U)] > \mathbb{E}_{G_\U}[PC_{mn}(\U,\V_2; G_\U)]$, if $p > q \frac{|P_{\V_1} \backslash P_\U|}{|P_\U \backslash P_{\V_2}|} $.

\end{itemize}
\label{thm}
\end{theoremx}

{\it proof} 
We use the same notation as for the proof of Lemma~\ref{lem}. 
\begin{itemize}
\item[(i)] Follows directly from the condition $|P_\U|^2 < |P_{\V_1}|\cdot|P_{\V_2}|$.

\item[(ii)] In the previous proof, we showed that $$\mathbb{E}[PC_{mn}(\U,\V_1;G_\U)] \geq Z_1 \mbox{ and }Z_2 \geq \mathbb{E}[PC_{mn}(\U,\V_2;G_\U)],$$ where $Z_1 = \frac{pa}{pa+1/2qx_1}$ and $Z_2 = \frac{a-x_2}{a-1/2x_2}$. We only need to show that $Z_1 > Z_2$ whenever $px_2 > qx_1$: 

$$
Z_1 = \frac{pa}{pa+1/2qx_1} \\
> \frac{pa}{pa+1/2px_2} \\
> \frac{a - x_2}{(a+1/2x_2) - x_2} = Z_2.
$$
\end{itemize}

The conditions required in the theorem above are easily satisfied when $\V_1$ and $\V_2$ are perturbations of $\U$. One of them states that the coarsening perturbation of $\U$ must be {\it as important as} the refinement perturbation: the geometric mean between $|P_{\V_1}|$  and $|P_{\V_2}|$ must be greater than $|P_{\U}|$. The second condition requires the ratio between $p$ and $q$ ---the intra and inter-edge densities--- to be larger as $\V_1$ gets coarser. 
A consequence of this result is that
none of the two measures $PC_{mn}(\U,\V)$, $PC_{mn}(\U,\V; G)$ directly captures how `close' partition $\V$ is to the ground truth partition $\U$. Instead,
$PC_{mn}(\U,\V)$ measures how close $\V$ is to being a {\it refinement} of $\U$, whereas $PC_{mn}(\U,\V; G)$ measures the opposite, how close $\V$ is to being a {\it coarsening} of $\U$. For this reason, when used together, the graph-aware and graph-agnostic measures give indications on the containment of parts of $\V$ and $\U$. Getting high values with respect to both measures indicates that the partitions are indeed similar.  

\subsection{Adjusting the graph-aware measures}

The expected value of the graph-aware measures of two independent partitions does not take a constant value. In fact, the expectation depends on the graph topology. Here, we propose an adjustment that does not depend on the graph topology but that considerably reduces the baseline, i.e., the expected similarity of random partitions. Recall that an adjusted similarity measure is obtained from a similarity measure by subtracting the expected value and re-normalizing properly:
$$AdjSim(\U, \V) = \frac{Sim(\U,\V) - \mathbb{E}\big[~Sim(\U,\V)~\big]}{1 - \mathbb{E}\big[~Sim(\U,\V)~\big]},$$
where the expectation is computed over all partitions $\U, \V$ from some random model. The permutation model used as the random model to adjust pair-counting measures on the set $V$ is not suitable here as it does not yield connected graph partitions. A simple random model that can be used for graph partitions assumes constant values of the number of internal edges: two partitions are drawn randomly with a fixed number of internal edges $|b_\U|$ and $|b_\V|$ respectively. We call this model the {\it fix-intra-edges random model}. The rough approximation we make under this model, is the following:
$$\mathbb{E}\big[~|b_\U \cdot b_\V| ~:~ |b_\U|, |b_\V|, G~\big] \approx \frac{|b_\U| \cdot |b_\V|}{|E|}.$$

From the approximation above, adjustments to each graph-aware measures defined in Table~\ref{tab:gpc} can be obtained. We omit the computation details, and we give the resulting adjusted measures:
\begin{eqnarray*}
APC_{f}(\U, \V;G) &=& \frac{|b_\U \cdot b_\V|  - \frac{|b_\U|\cdot|b_\V|}{|E|}}{f(|b_\U|, |b_\V|) - \frac{|b_\U|\cdot|b_\V|}{|E|}}.
\end{eqnarray*}
Just as it is the case for set measures, the adjusted graph-aware Rand Index ($ARI(\cdot;G)$) is the same as one of the adjusted graph-aware pair counting measures:
$$\mathit{ARI}(\U,\V;G) = APC_{mn}(\U, \V;G) = \frac{|b_\U \cdot b_\V|  - \frac{|b_\U|\cdot|b_\V|}{|E|}}{\frac{|b_\U|+|b_\V|}{2} - \frac{|b_\U|\cdot|b_\V|}{|E|}}.$$

\section{Experiments}
\label{section:both}
To test graph algorithms' reliability, different benchmarks have been developed to generate graphs and their associated ground truth partitions \cite{girvan2002community, condon2001algorithms, decelle2011asymptotic, lfr, lancichinetti2009benchmarks}. The LFR model \cite{lfr, lancichinetti2009benchmarks} was designed to reproduce certain topological properties observed in real-world networks: the size of the communities is power-law distributed, and so is the node degree. A typical way to assess the superiority of a partitioning algorithm over another is to use a family of LFR  graphs that range from clear partition structure (low inter-part edge density) to practically no partition structure (high inter-part edge density) and to plot the similarities between the output partitions and the true graph partitions of this family against the inter-part edge density $\mu$. This produces a {\it similarity curve} for each algorithm and conclusions are made based on those curves: higher curves imply better algorithms \cite{yang2016comparative}. 

In this section, we study the impact of the adjustment on the graph-aware measures and we illustrate the usefulness of Theorem~\ref{thm} on data, i.e., the complementarity of graph-aware and graph-agnostic measures.

\subsection{Adjusted graph-aware measures}
\label{sec:generation}

To study the expected value of the adjusted graph-aware measures on random partitions, we need to generate random connected partitions of graphs. We use two different generation processes.

\noindent{\bf Generation Process 1:} Fix $k$, the size of the partition. From a random vertex, generate a depth-first search tree that spans $G$ and delete $k-1$ random edges from the tree. The remaining $n-k$ edges of the tree are considered class-1 edges, yielding a binary vector $b$ where $|b|=n-k$. We then get its associated connected graph partition $U$ using its class representative: $b_\U = \overline{b}^{_G}$. 

\begin{figure}[ht]
\begin{center}
\centering
\subfloat[]{
\includegraphics[angle=0, width=.5\linewidth]{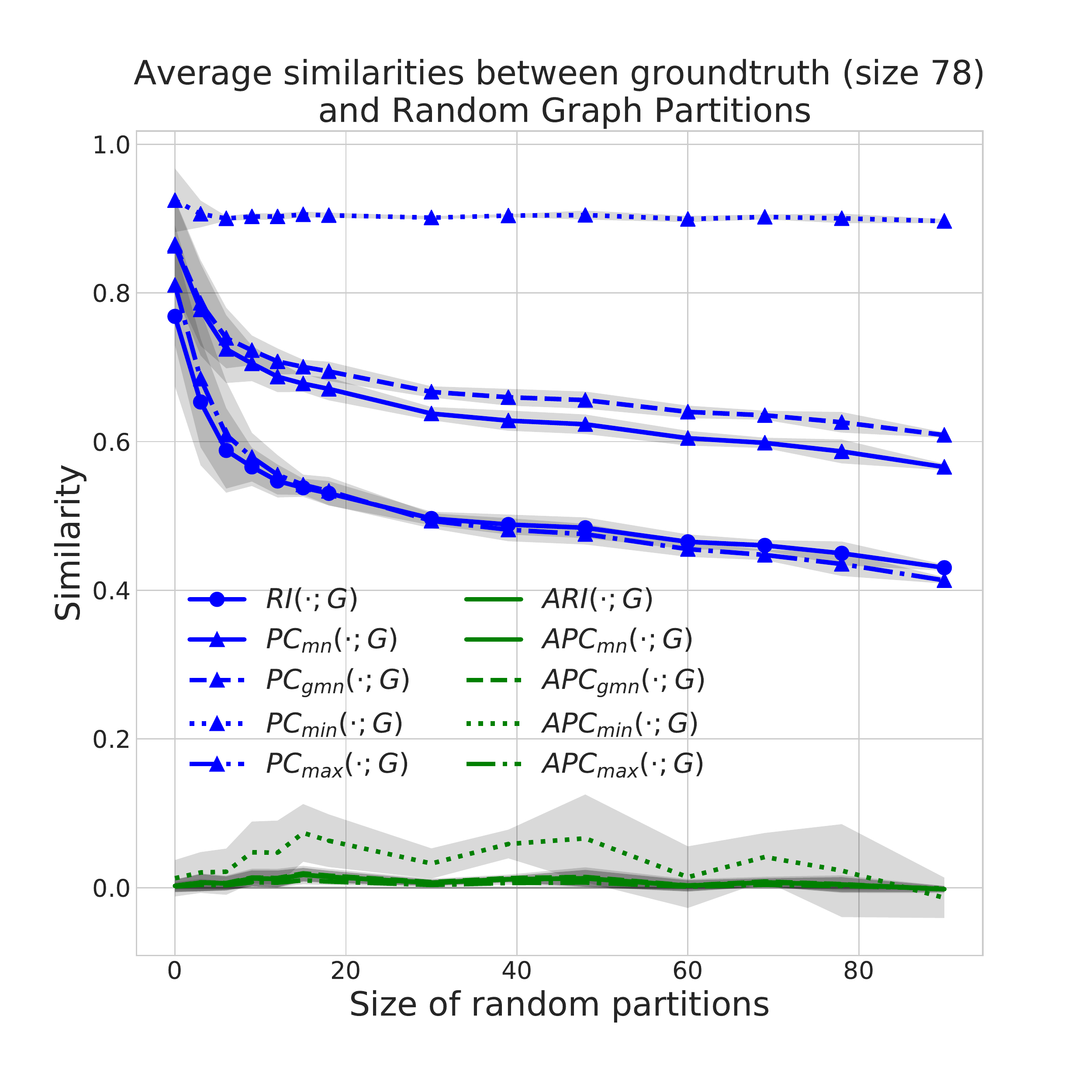}}
\subfloat[]{
\includegraphics[angle=0, width=.5\linewidth]{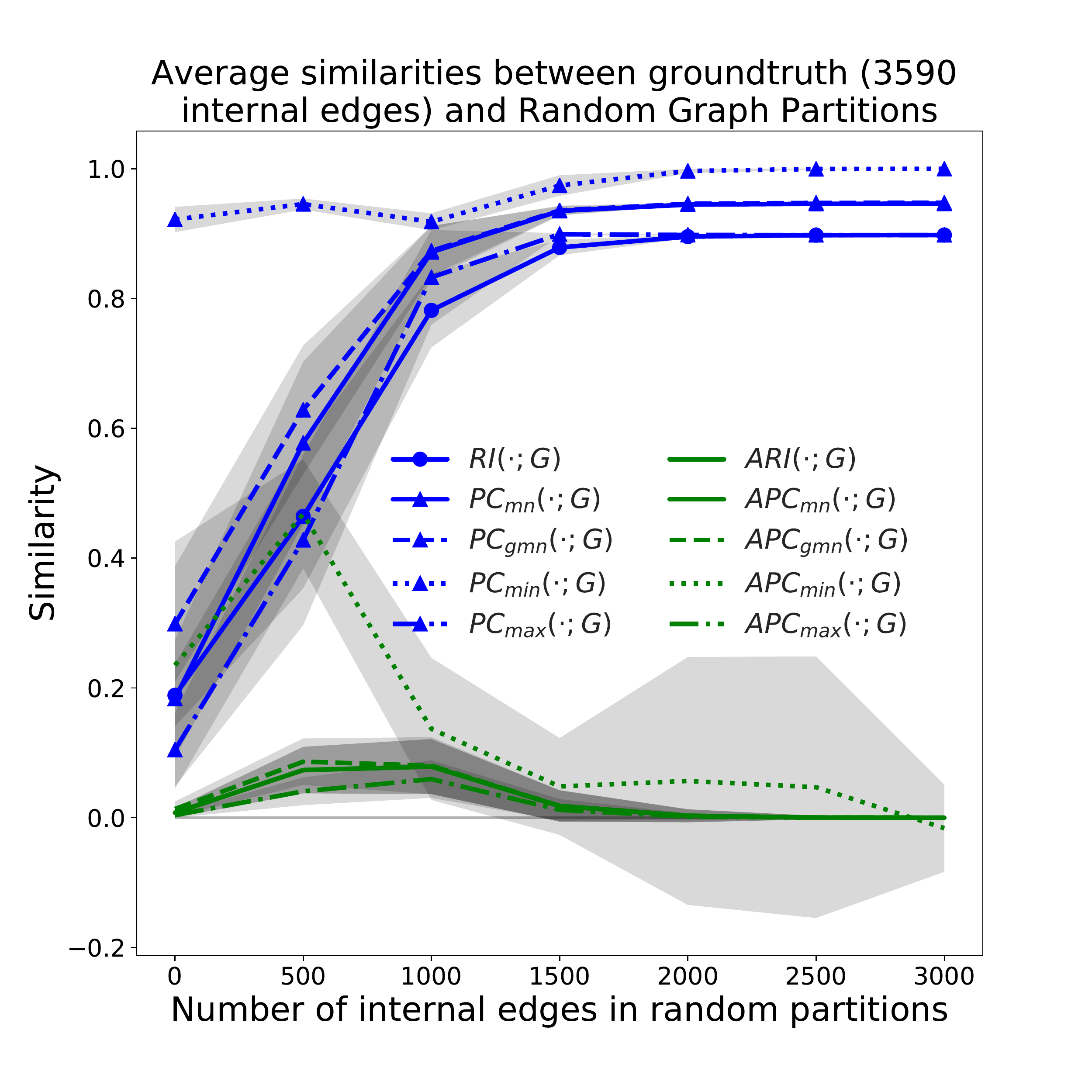}}
\caption{\small Similarity measurements between a ground truth graph partition of size 78 and random partitions having (a) pre-determined sizes, (b) pre-determined number of internal edges. $|E(G)|=4000$, $\mu=0.1$ which yields a ground truth partition with $3590$ internal edges. The measures were smoothed on windows of size 5 in (a) on size 250 in (b), shaded regions indicate the standard deviations.} 
\label{fig:fix_internal}
\end{center}
\end{figure}

In Figure~\ref{fig:fix_internal} (a), we show the adjusted and non-adjusted similarity measures between ground truth partitions obtained from the LFR model\footnote{The generation parameters are given in Appendix~\ref{app:param}.} and random partitions generated according to Process 1. As we can see, the adjusted measures are much closer to a 0-baseline for independent partitions. Only four curves are shown as adjusted functions since $ARI(\cdot;G)$ and $APC_{mn}(\cdot;G)$ collapse to the same function. We see that the adjusted measure $APC_{min}(\cdot;G)$ has much higher baseline and variance compared to the others. 

\noindent{\bf Generation Process 2:} The second random graph partition generation consists in randomly selecting $k$ edges of $G$ as class-1 edges, i.e., randomly select a binary vector in $\{b \in \{0,1\}^{|E|} : |b|=k\}$. Then, we get its associated connected graph partition $U$ using its class representative: $b_\U = \overline{b}^{_G}$. 

In Figure~\ref{fig:fix_internal} (b), we see the similarities between a graph's communities and some random partitions containing a pre-determined number of internal edges. From the plot, we see that the adjustment reduces the baseline of the measures considerably except for the $APC_{min}(\cdot;G)$. However, this plot demonstrates that the approximation we are using,  
$\mathbb{E}\left[|b_\U \cap b_\V|~:~|b_\V|\right] = |b_\V|\cdot|b_\U|/|E|$, for a fix $b_\U$, is wrong. 
The quantity $|b_\U| \cdot |b_\V|/|E|$ is a good estimation for graphs with no community structure: trees, complete graphs or Erd\"os-Renyi random graphs. For graphs with community structure, the approximation underestimates the true expectation so the adjusted measures are still above the desired 0-baseline.
\begin{figure}[ht]
\begin{center}
\centering
\includegraphics[angle=0, width=.7\linewidth]{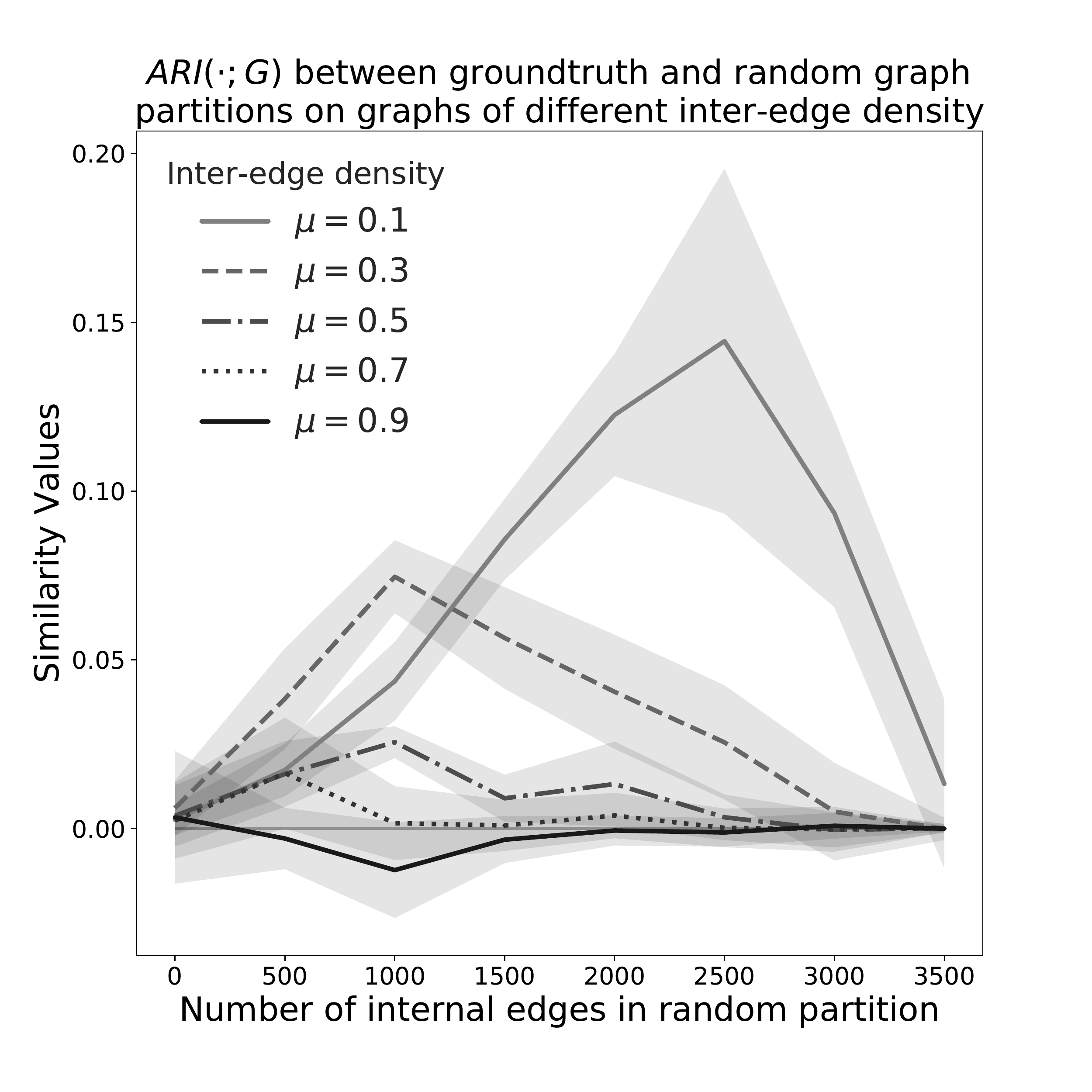}
\caption{\small Graph-aware adjusted Rand Index between random partitions and ground truth partition for graphs $G_\mu$ with various community structure strengths (small $\mu$ implies strong community structure). 10 independent graphs were generated for each $\mu$-value,  shaded regions indicate the standard deviations.} 
\label{fig:various}
\end{center}
\end{figure}
This statement is illustrated in Figure~\ref{fig:various}. The similarity measure $ARI(\cdot;G)$ is computed between random partitions and the ground truth partitions of LFR graphs having various level of community structure: low $\mu$-values indicate low inter-partition edge density, so strong community structure. As one can see, for graphs with strong community structure, the adjustment estimation is worse than on weak community structure graphs. 
The issue is that a good estimation of $\mathbb{E}_\V(~|b_\U \cap b_\V|~:~|b_\V|)$ given $G$ and $\U$ is still an open question.

\subsection{Adjusted graph-aware vs. graph-agnostic measures}
\label{sec:exp_comparison}
\begin{figure}[ht]
\begin{center}
\centering
\includegraphics[angle=0, width=1.1\linewidth]{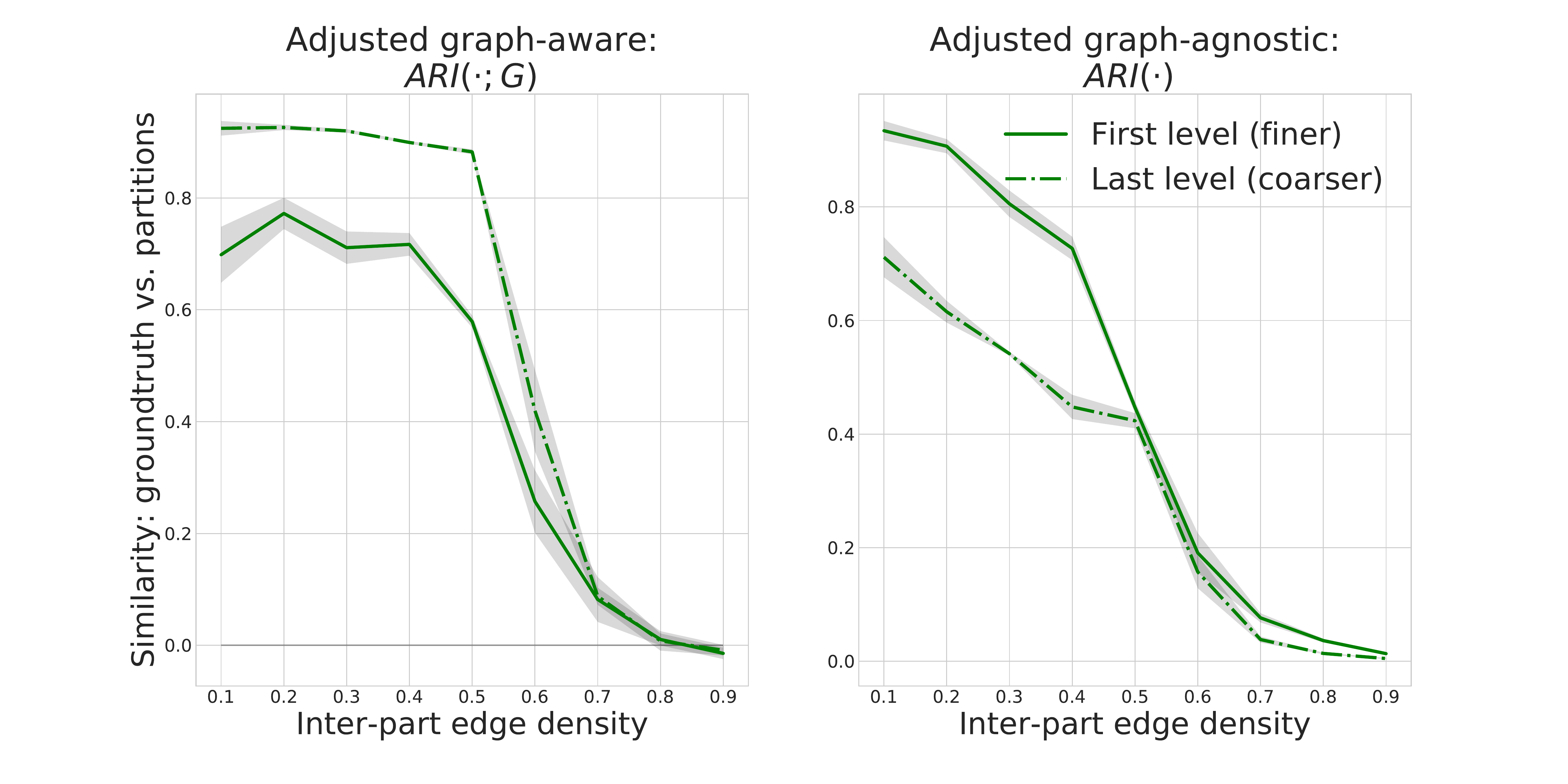}
\caption{\small Comparing the similarity curves of a partition (coarser) and a refinement of it (finer) of LFR graphs having varying inter-cluster edge densities. The graph-aware and graph-agnostic measures yield contradicting conclusions. 10 independent graphs were generated for each $\mu$-value, shaded regions indicate the standard deviations.} 
\label{fig:resolution}
\end{center}
\end{figure}

We limit our comparisons to the graph-aware and agnostic variants of the Rand Index  and the adjusted mutual information. Theorem~\ref{thm} shows that the unadjusted versions of the measures penalize refinements and coarsenings in opposite ways. In Figure~\ref{fig:resolution}, we present empirical evidences that the same is true for the adjusted versions of the measures.
We compare the similarity curves of two graph partition algorithms: the first level and the last level of a hierarchy of partitions obtained with the Louvain method \cite{louvain}. The algorithms compared correspond to two different resolutions, one being a refinement of the other. As one can see, the graph-agnostic and graph-aware measures yield contradicting conclusions. According to the graph-agnostic measures $ARI$, the finer partitions are more similar to the ground truth partitions, therefore, a better choice of algorithm. Note that we obtain the same conclusion when using $AMI$. When using the graph-aware measure $ARI(\cdot,G)$, the conclusion is the opposite: the coarser partitions are closer to the ground truth partitions. This is a good illustration of the fact that graph-agnostic similarities measure how close a partition is to being a {\it refinement} of the ground truth partition while graph-aware captures how close a partition is to being a {\it coarsening} of the ground truth partition. It is therefore not possible to assess the superiority of any of the two algorithms compared in Figure~\ref{fig:resolution} when using both types of measures: one produces a refinement and the other a coarsening of the ground truth partition.

\subsection{Impacts on Comparison Study Conclusion}

In 2017, \cite{yang2016comparative} provided an exhaustive comparison study of graph clustering algorithms on artificial LFR networks. The families of graphs generated for the study are strongly structured, hence the large majority of their conclusions hold regardless of the choice of measure: adjusted or not, graph-aware or not. 
However, in cases where the sizes of the partitions differ significantly, where one of the algorithm underestimates while the other overestimates the number of clusters with respect to the ground truth partition, the choice of measure does have an impact. 
\begin{figure}[ht]
\begin{center}
\centering
\includegraphics[angle=0, width=.62\linewidth]{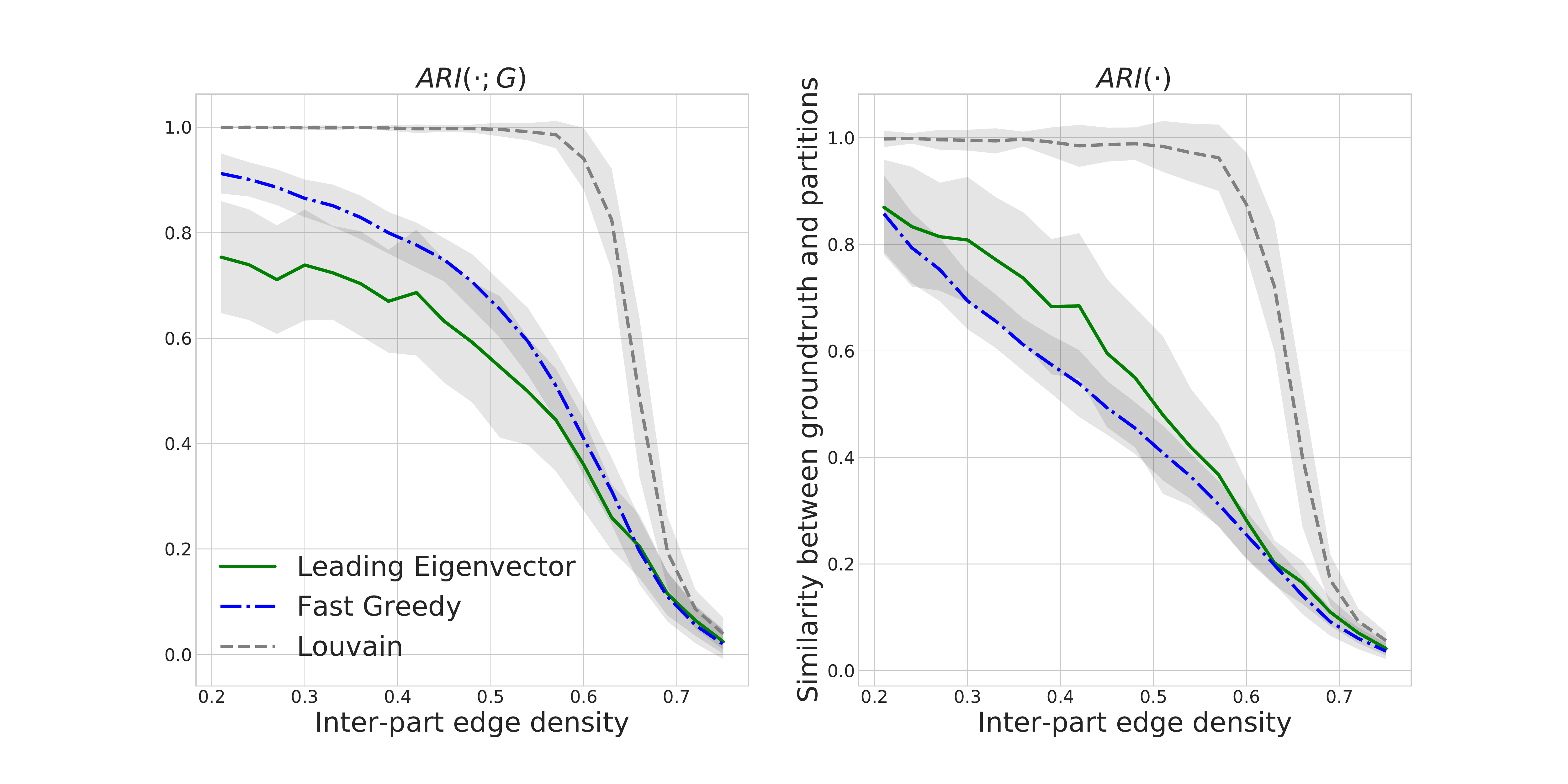}%
\includegraphics[angle=0, width=.3\linewidth]{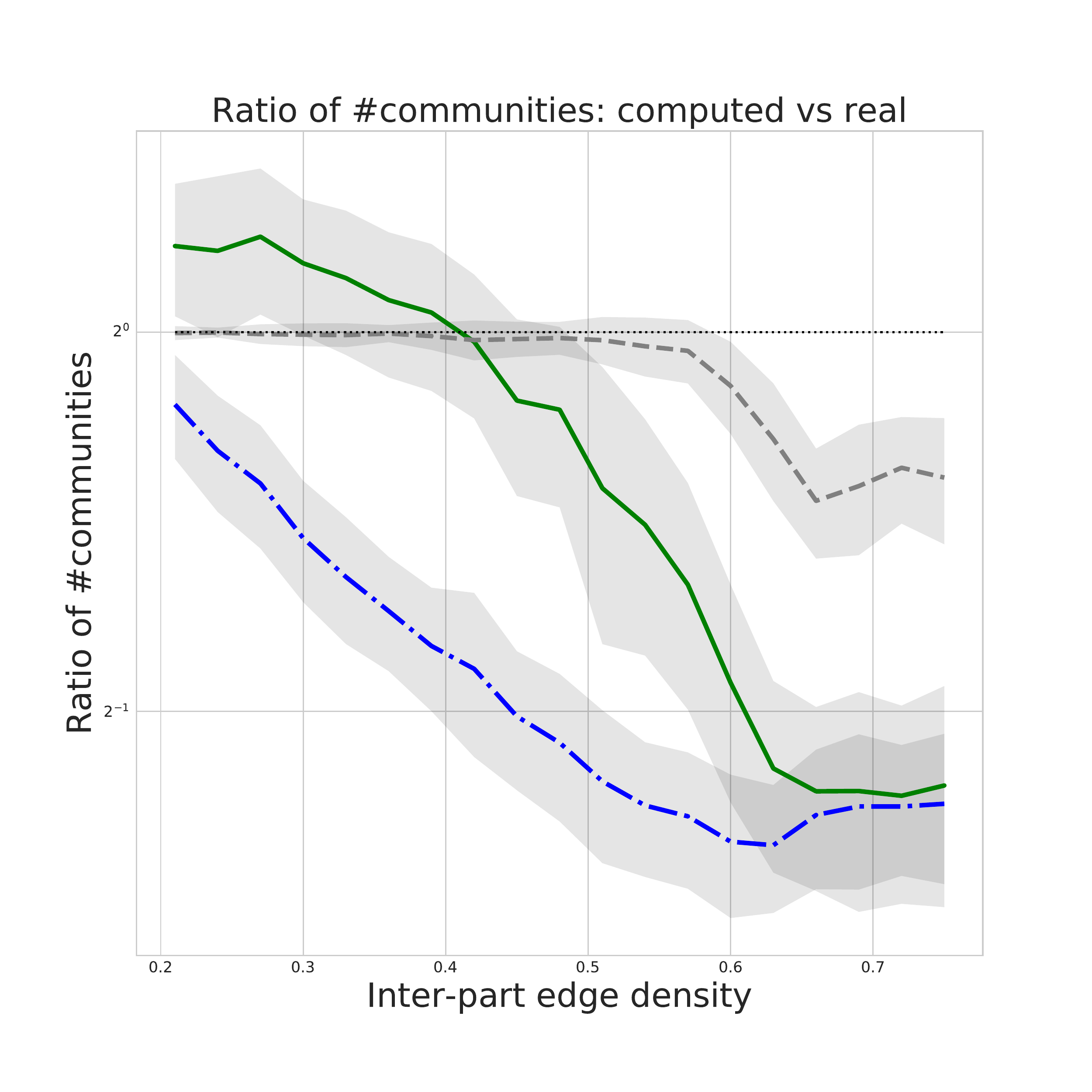}
\caption{\small Comparing the similarity curves of three partition algorithms on LFR graphs. The graph-aware and graph-agnostic have contradicting conclusions for the algorithms Leading Eigenvector and Fast Greedy. The right plot indicates that Leading Eigenvector produces finer partitions than Fast Greedy. The Louvain algorithm outperforms the other two. 100 independent graphs were generated for each $\mu$-value, shaded regions indicate the standard deviations.} 
\label{fig:study2017}
\end{center}
\end{figure}

In Figure~\ref{fig:study2017}, we show the similarity curves of the {\it FastGreedy algorithm} \cite{clauset2004finding}, the {\it Leading eigenvector algorithm} \cite{newman2006finding} and the {\it Louvain method} on one of the LFR family used in the study\footnote{The LFR parameters are given in Appendix~\ref{app:param}.}. The choice of measure in that case impacts the conclusion: the graph-agnostic measure shows that FastGreedy outperforms Leading Eigenvector, whereas the graph-aware measure shows the opposite. This can partly be explained by the fact that Fast Greedy underestimates the number of clusters, as shown in Figure~\ref{fig:study2017} (bottom plot), and therefore produces better coarsenings of the ground truth partitions than Leading Eigenvector. Again, according to this experiment, it is not possible to claim which of the two algorithms has higher accuracy. However, it is possible to claim that the Louvain method outperforms the other two algorithms on this family of graphs since it has higher similarity curves with respect to both measures.

\section{Conclusion}
\label{section:conclusion}
\noindent In this paper, we proposed an adjusted form of graph-aware similarity measures for graph partitions. We demonstrated that the graph-aware measures and graph-agnostic measures are complementary in that they behave differently with respect to refinements and coarsenings of partitions. As a consequence, both measures should be used jointly to assess similarity of graph partitions, using a single measure can lead to wrong conclusions in the study of graph partitioning algorithms.

The estimation used for the adjustment of the graph-aware measures is not tight for structured graphs. Hence, finding a better estimation of the adjustment under the fix-intra-edges random model that incorporates information about the graph's topology would increase the adjusted graph-aware measures' reliability. This is a topic for future work.

\appendix
\section{Adjustment for graph-aware measures}
\label{app:adj}
To compute the adjustment of the graph-aware measures, note that $f(|b_\U|, |b_\V|)$ is a constant under the conditional expectation so we have $$\mathbb{E}\left[ PC_f(\U,\V;G) : |b_\U|, |b_\V|,G\right] = \frac{|b_\U| \cdot |b_\V|}{|E| \cdot f(|b_\U|, |b_\V|)}.$$ 
For the Rand Index, since $|(1-b_\U) \cdot (1-b_\V)| = |E| - (|b_\U|+|b_\V|) + |b_\U \cdot b_\V|$, we can write $RI(\U,\V, G)=1 -\frac{|b_\U| + |b_\V|}{|E|}+ \frac{2 |b_\U \cdot b_\V|}{|E|}$ and so $$\mathbb{E}\left[ RI(\U,\V;G) : |b_\U|, |b_\V|,G\right] = 1 -\frac{|b_\U| + |b_\V|}{|E|}+ \frac{2 |b_\U| \cdot |b_\V|}{|E|^2}.$$ The rest is obtained using the definition $$\frac{Sim(\U,\V) - \mathbb{E}[Sim(\U,\V)]}{1 - \mathbb{E}[Sim(\U,\V)]}.$$

\vspace{.1in}
\section{Graph generation parameters}
\label{app:param}
Experiments on Figures 3, 4, 5 and 6 were obtained by generating LFR graphs using the first set of parameters of Table~\ref{tab:param}, while the graph generation for Figure 7 used the second set of parameters of the table. {\it Auto} indicates that the value is automatically obtained by the generation algorithm. 
\begin{table}[h]
\caption{\small Graph generation parameters for LFR algorithm.}
\label{tab:param}
\begin{center}
\begin{tabular}{l|llllll}
& \multicolumn{3}{c}{Degree} & \multicolumn{3}{c}{Community sizes} \\
N& Max & Avg & Exp. & Min & Max & Exp. \\
\hline
1000 & 8 & 8 & -1 & 10 & 15 & -1 \\
233 & 23 & 20 & -2 & Auto & 23 & -1
\end{tabular}
\end{center}
\end{table}

\bibliographystyle{plain}
\bibliography{references}

\begin{thebibliography}{10}

\bibitem{lfr}
S.~Fortunato A.~Lancichinetti and F.~Radicchi.
\newblock {Benchmark graphs for testing community detection algorithms}.
\newblock {\em Phys. Rev. E}, 78(046110), 2008.

\bibitem{Albatineh2006}
Ahmed~N. Albatineh, Magdalena Niewiadomska-Bugaj, and Daniel Mihalko.
\newblock On similarity indices and correction for chance agreement.
\newblock {\em Journal of Classification}, 23(2):301--313, Sep 2006.

\bibitem{Cha}
S-S Choi, S-H Cha, and C.~Tappert.
\newblock {A Survey of Binary Similarity and Distance Measures}.
\newblock {\em J. Systemics, Cybernetics and Informatics}, (8), 2010.

\bibitem{clauset2004finding}
Aaron Clauset, Mark~EJ Newman, and Cristopher Moore.
\newblock Finding community structure in very large networks.
\newblock {\em Physical review E}, 70(6):066111, 2004.

\bibitem{condon2001algorithms}
Anne Condon and Richard~M Karp.
\newblock Algorithms for graph partitioning on the planted partition model.
\newblock {\em Random Structures and Algorithms}, 18(2):116--140, 2001.

\bibitem{decelle2011asymptotic}
Aurelien Decelle, Florent Krzakala, Cristopher Moore, and Lenka Zdeborov{\'a}.
\newblock Asymptotic analysis of the stochastic block model for modular
  networks and its algorithmic applications.
\newblock {\em Physical Review E}, 84(6):066106, 2011.

\bibitem{Fortunato36}
Santo Fortunato and Marc Barth{\'e}lemy.
\newblock Resolution limit in community detection.
\newblock {\em Proceedings of the National Academy of Sciences}, 104(1):36--41,
  2007.

\bibitem{fortunato2016community}
Santo Fortunato and Darko Hric.
\newblock Community detection in networks: A user guide.
\newblock {\em Physics Reports}, 659:1--44, 2016.

\bibitem{gates2017impact}
Alexander~J Gates and Yong-Yeol Ahn.
\newblock The impact of random models on clustering similarity.
\newblock {\em The Journal of Machine Learning Research}, 18(1):3049--3076,
  2017.

\bibitem{girvan2002community}
Michelle Girvan and Mark~EJ Newman.
\newblock Community structure in social and biological networks.
\newblock {\em Proceedings of the national academy of sciences},
  99(12):7821--7826, 2002.

\bibitem{arabie}
L.~Hubert and P.~Arabie.
\newblock {Comparing partitions}.
\newblock {\em Journal of Classification}, (193-218), 1985.

\bibitem{Kumpula2007}
J.~M. Kumpula, J.~Saram{\"a}ki, K.~Kaski, and J.~Kert{\'e}sz.
\newblock Limited resolution in complex network community detection with potts
  model approach.
\newblock {\em The European Physical Journal B}, 56(1):41--45, 2007.

\bibitem{lancichinetti2009community}
A~Lancichinetti and S~Fortunato.
\newblock Community detection algorithms: a comparative analysis.
\newblock {\em Physical review. E, Statistical, nonlinear, and soft matter
  physics}, 80(5 Pt 2):056117, 2009.

\bibitem{lancichinetti2009benchmarks}
Andrea Lancichinetti and Santo Fortunato.
\newblock Benchmarks for testing community detection algorithms on directed and
  weighted graphs with overlapping communities.
\newblock {\em Physical Review E}, 80(1):016118, 2009.

\bibitem{meila2005}
M.~Meil\u{a}.
\newblock {Comparing clusterings - An Axiomatic View}.
\newblock {\em Proceedings of the 22nd International Conference on Machine
  Learning}, 2005.

\bibitem{meila}
M.~Meil\u{a}.
\newblock {Comparing clusterings - an information based distance}.
\newblock {\em Journal of Multivariate Analysis}, (98), 2007.

\bibitem{newman2006finding}
Mark~EJ Newman.
\newblock Finding community structure in networks using the eigenvectors of
  matrices.
\newblock {\em Physical review E}, 74(3):036104, 2006.

\bibitem{orman2009comparison}
G{\"u}nce~Keziban Orman and Vincent Labatut.
\newblock A comparison of community detection algorithms on artificial
  networks.
\newblock In {\em International Conference on Discovery Science}, pages
  242--256. Springer, 2009.

\bibitem{raghavan2007near}
Usha~Nandini Raghavan, R{\'e}ka Albert, and Soundar Kumara.
\newblock Near linear time algorithm to detect community structures in
  large-scale networks.
\newblock {\em Physical review E}, 76(3):036106, 2007.

\bibitem{rand1971}
William~M. Rand.
\newblock Objective criteria for the evaluation of clustering methods.
\newblock {\em Journal of the American Statistical Association},
  66(336):846--850, 1971.

\bibitem{reichardt2006statistical}
J{\"o}rg Reichardt and Stefan Bornholdt.
\newblock Statistical mechanics of community detection.
\newblock {\em Physical Review E}, 74(1):016110, 2006.

\bibitem{romano2016adjusting}
Simone Romano, Nguyen~Xuan Vinh, James Bailey, and Karin Verspoor.
\newblock Adjusting for chance clustering comparison measures.
\newblock {\em The Journal of Machine Learning Research}, 17(1):4635--4666,
  2016.

\bibitem{rosvall2007information}
Martin Rosvall and Carl~T Bergstrom.
\newblock An information-theoretic framework for resolving community structure
  in complex networks.
\newblock {\em Proceedings of the National Academy of Sciences},
  104(18):7327--7331, 2007.

\bibitem{lapalme}
M.~Sokolova and G.~Lapalme.
\newblock {A systematic Analysis of Performance Measures for Classification
  Tasks}.
\newblock {\em Information Processing and Management}, (45), 2009.

\bibitem{louvain}
R.~Lambiotte V.D.~Blondel, J.-L.~Guillaume and E.~Lefebvre.
\newblock {Fast unfolding of communities in large networks}.
\newblock {\em J. Stat. Mech.}, (P10008), 2008.

\bibitem{vinh}
N.~X. Vinh, J.~Epps, and J.~Bailey.
\newblock {Information Theoretic Measures for Clusterings Comparison: Is a
  Correction for Chance Necessary?}
\newblock {\em Proceedings of the 26th International Conference on Machine
  Learning}, 2009.

\bibitem{vinh2010}
N.~X. Vinh, J.~Epps, and J.~Bailey.
\newblock {Information Theoretic Measures for Clusterings Comparison: Variants,
  Properties, Normalization and Correction for Chance}.
\newblock {\em Journal of Machine Learning Research}, (11), 2010.

\bibitem{yang2016comparative}
Zhao Yang, Ren{\'e} Algesheimer, and Claudio~J Tessone.
\newblock A comparative analysis of community detection algorithms on
  artificial networks.
\newblock {\em Scientific Reports}, 6:30750, 2016.

\end{thebibliography}

\end{document}